\documentclass[letterpaper, 10 pt, conference]{ieeeconf}  

\IEEEoverridecommandlockouts                              

\input{common.sty}
\usepackage{flushend}

\let\labelindent\relax
\usepackage[shortlabels]{enumitem}

\title{\LARGE \bf
Contact Reduction with Bounded Stiffness for Robust Sim-to-Real Transfer of Robot Assembly
}

\author{Nghia Vuong$^{1}$, and Quang-Cuong
  Pham$^{1,2}$
  \thanks{$^{1}$Singapore Centre for 3D Printing (SC3DP), School of
    Mechanical and Aerospace Engineering, NTU, Singapore}
  \thanks{$^{2}$Eureka Robotics, Singapore}
}

\begin{document}

\maketitle
\thispagestyle{empty}
\pagestyle{empty}

\begin{abstract}
  In sim-to-real Reinforcement Learning (RL), a policy is trained in a simulated environment and then deployed on the physical system. The main challenge of sim-to-real RL is to overcome the \emph{reality gap} - the discrepancies between the real world and its simulated counterpart. Using general geometric representations, such as convex decomposition, triangular mesh, signed distance field can improve simulation fidelity, and thus potentially narrow the reality gap. Common to these approaches is that many contact points are generated for geometrically-complex objects, which slows down simulation and may cause numerical instability. Contact reduction methods address these issues by limiting the number of contact points, but the validity of these methods for sim-to-real RL has not been confirmed. In this paper, we present a contact reduction method with bounded stiffness to improve the simulation accuracy. Our experiments show that the proposed method critically enables training RL policy for a tight-clearance double pin insertion task and successfully deploying the policy on a rigid, position-controlled physical robot.
\end{abstract}


\section{INTRODUCTION} \label{sec-intro}

Learning robot manipulation skills through Reinforcement Learning (RL) is challenging. Modern RL algorithms typically have high sample complexities, resulting in lengthy robot execution time. Moreover, model-free RL algorithms perform random action sampling during the exploration phase, raising safety concerns when the robot interacts with its environment. One way to address these problems entails learning an RL policy in a virtual environment and deploying it on the physical system. Numerous studies have demonstrated the ability of this simulation-to-reality methodology to teach sophisticated robotics tasks such as dexterous in-hand manipulation \cite{andrychowiczLearningDexterous2020}, locomotion on unknown terrain. The main challenge of sim-to-real RL is overcoming the \emph{reality gap} - the discrepancies between the real world and its simulated counterpart. Simulated environments should be created to best represent the real world to reduce the reality gap. However, the majority of works build simulated environments using collision geometries that approximate the true geometry with primitive shapes (cylinders, boxes, spheres). For objects with concave surfaces, such a simple approximation may fail to realize the actual contact region.


A possible solution is to use geometric representations that can accurately represent complex surfaces. Examples of these geometric representations include convex decomposition, triangular mesh \cite{hauserRobustContact2016}, signed distance field \cite{narangFactoryFast2022}. Common to these approaches is that many contact points can be generated for geometrically-complex objects. Too many contact points reduce simulation speed and potentially cause numerical instability. To address this problem, several works have proposed contact reduction methods to limit the number of contact points \cite{hauserRobustContact2016}, \cite{otaduyModularHaptic2006}, \cite{narangFactoryFast2022}. However, none of these works explore how contact reduction methods influence simulation accuracy, which is crucial for sim-to-real reinforcement learning.

\begin{figure}[!t]
  \centering
  \subfloat[Case 1 - 3D view]{
    \includegraphics[width=0.3\columnwidth]{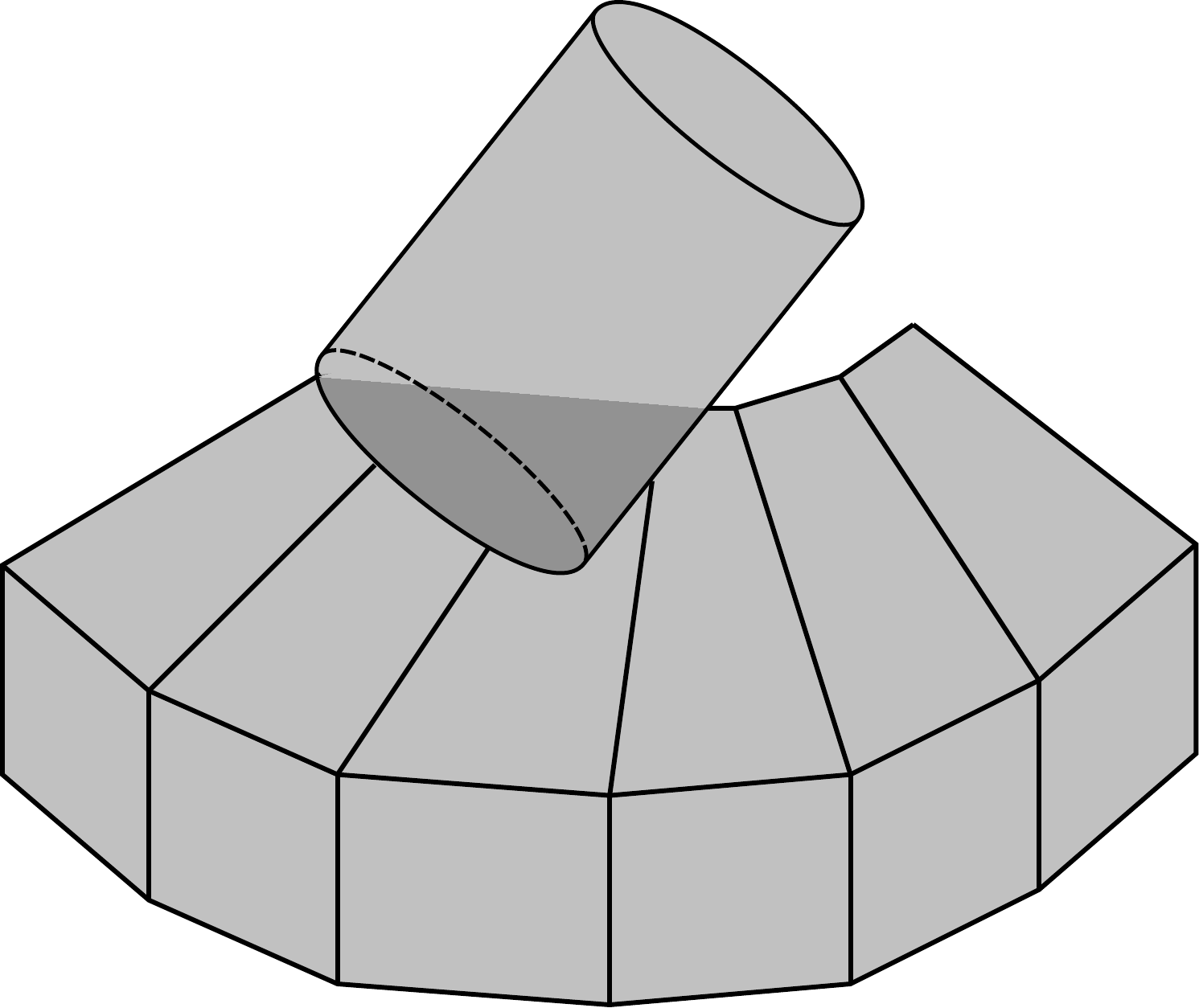}
  }\hspace{0.05\columnwidth}
  \subfloat[Case 1 - Front view]{
    \includegraphics[width=0.3\columnwidth]{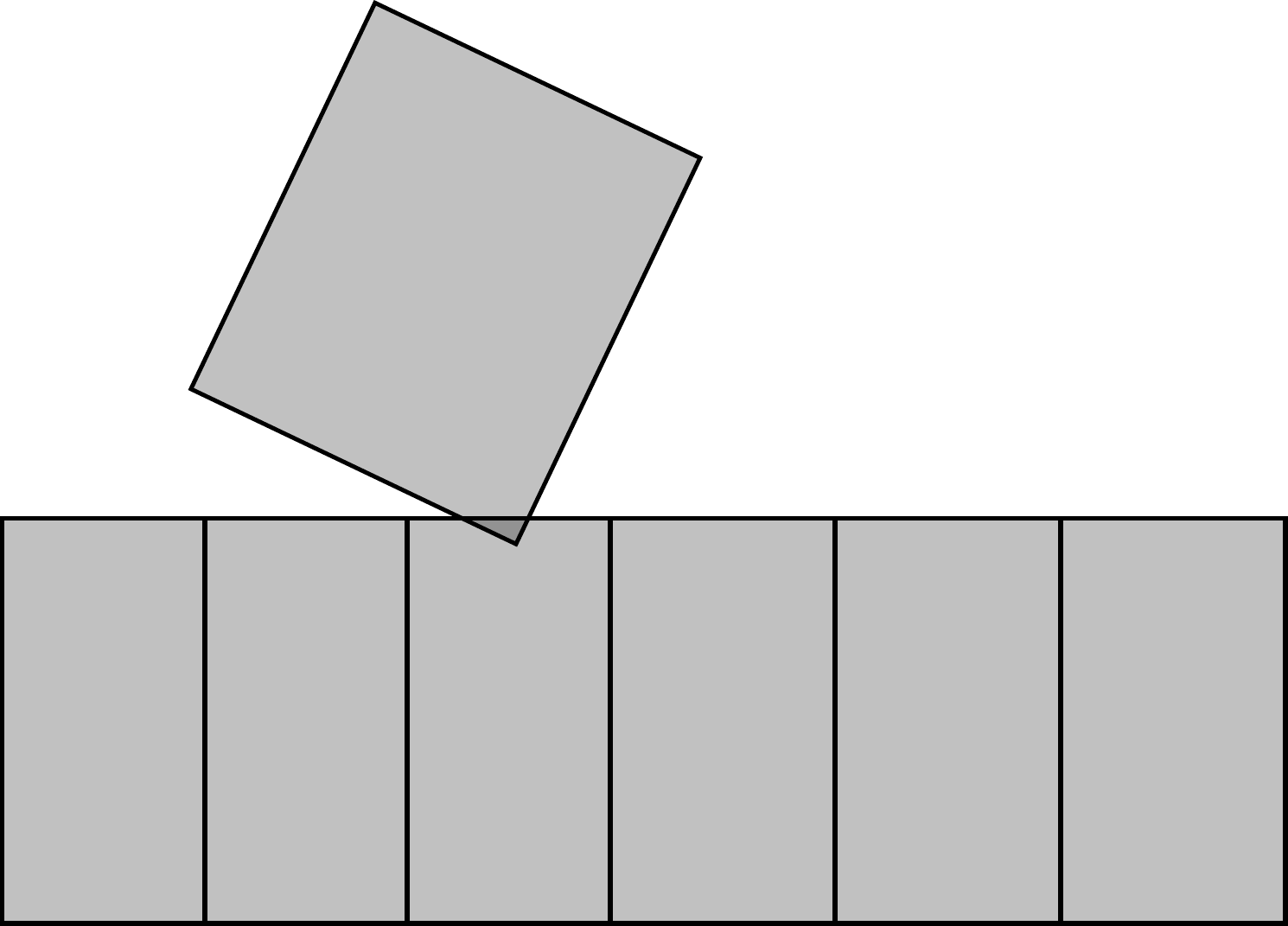}
  }\hspace{0.05\columnwidth}
  \subfloat[Case 1 - Interpretation as springs]{
    \includegraphics[width=0.15\columnwidth]{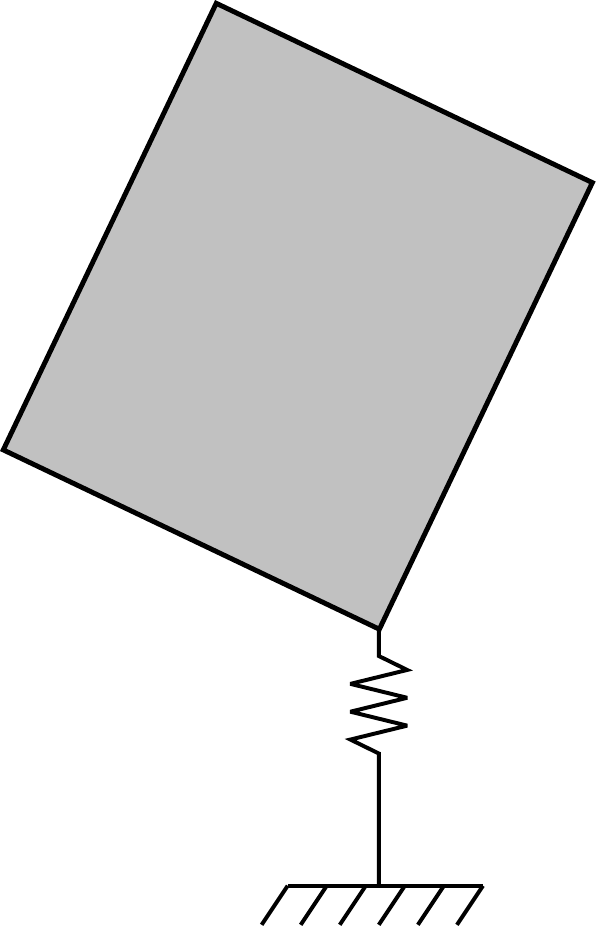}
  }\\
  \subfloat[Case 2 - 3D view]{
    \includegraphics[width=0.3\columnwidth]{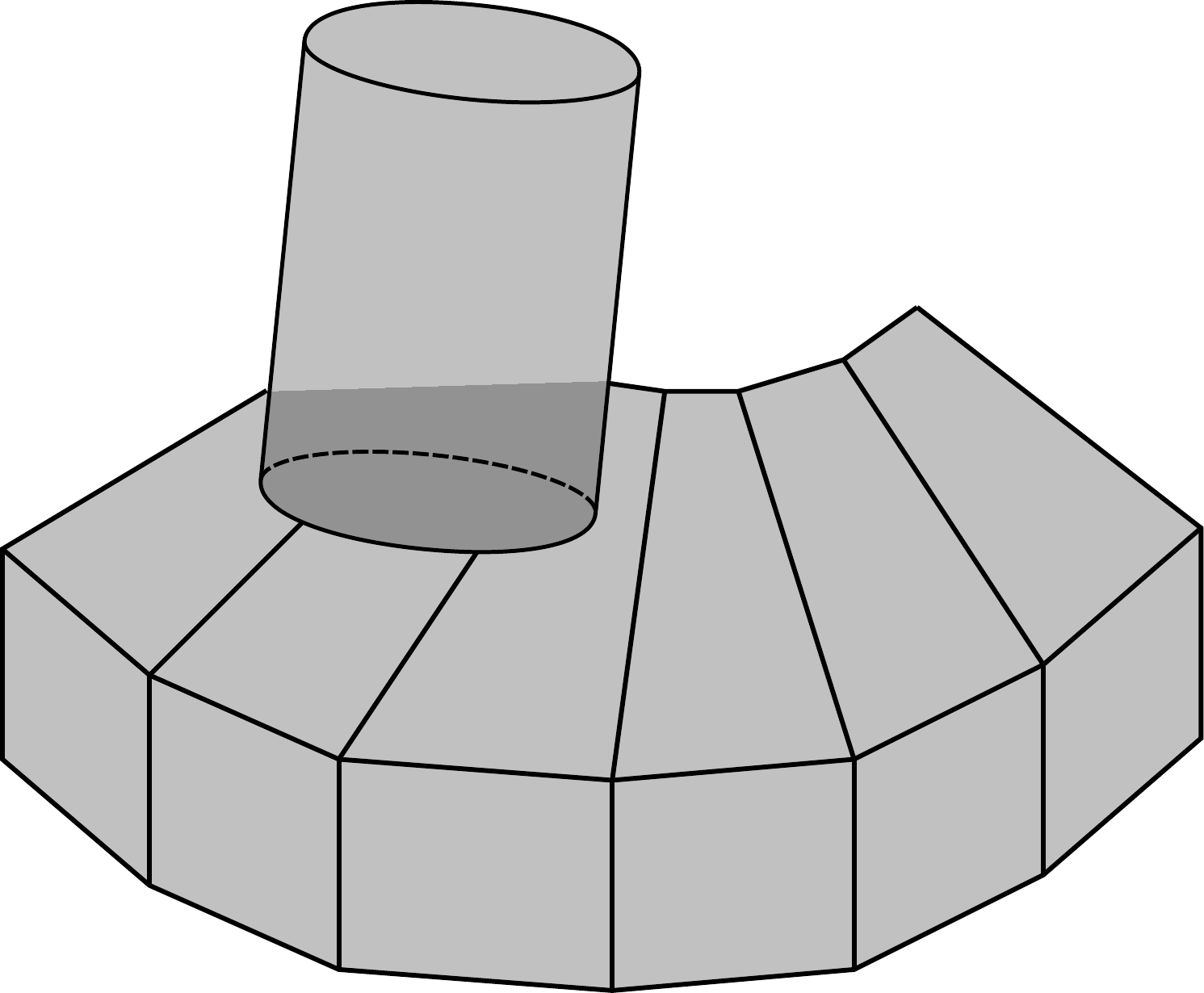}
  }\hspace{0.05\columnwidth}
  \subfloat[Case 2 - Front view]{
    \includegraphics[width=0.3\columnwidth]{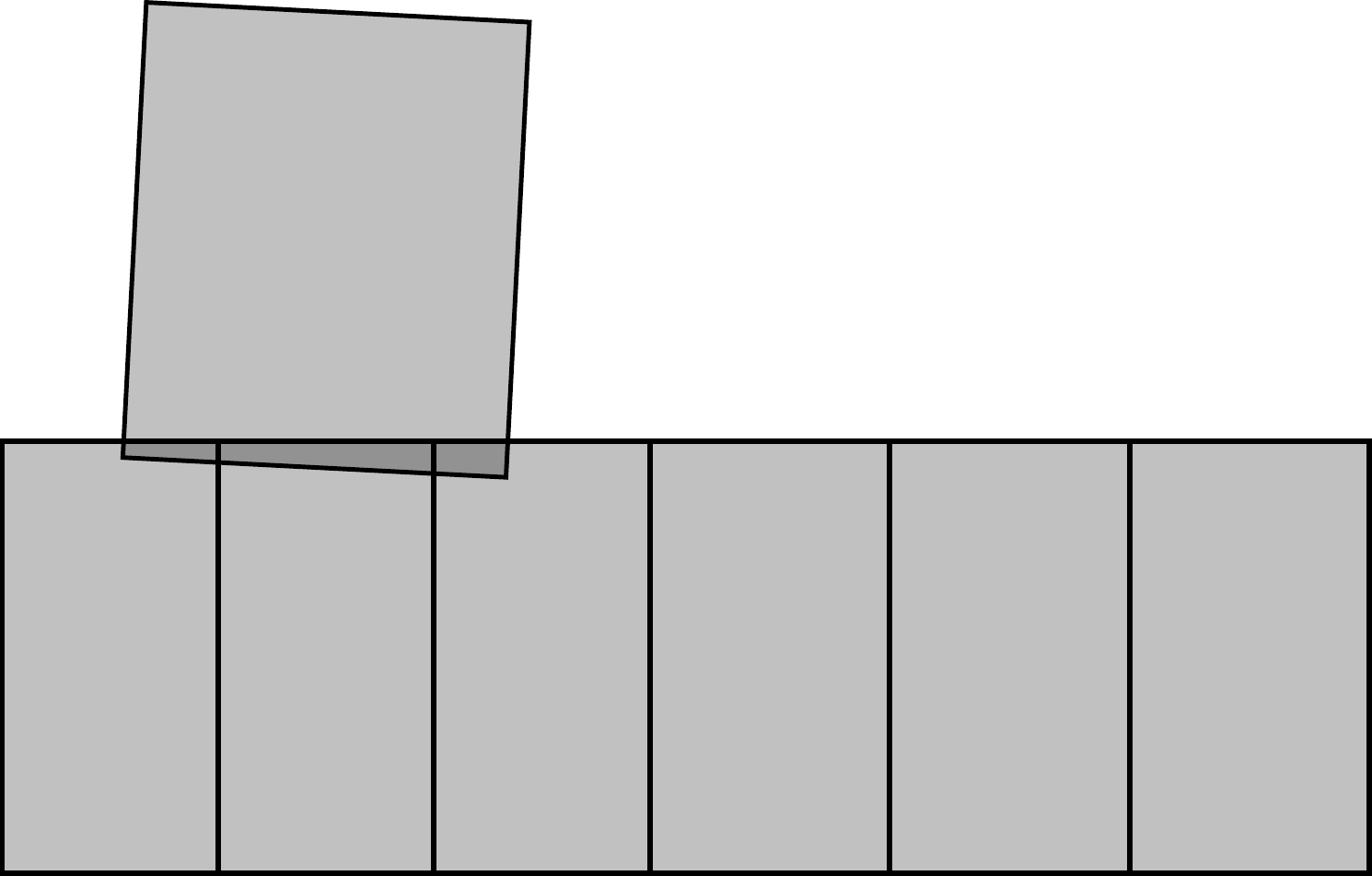}
  }\hspace{0.05\columnwidth}
  \subfloat[Case 2 - Interpretation as springs]{
    \includegraphics[width=0.15\columnwidth]{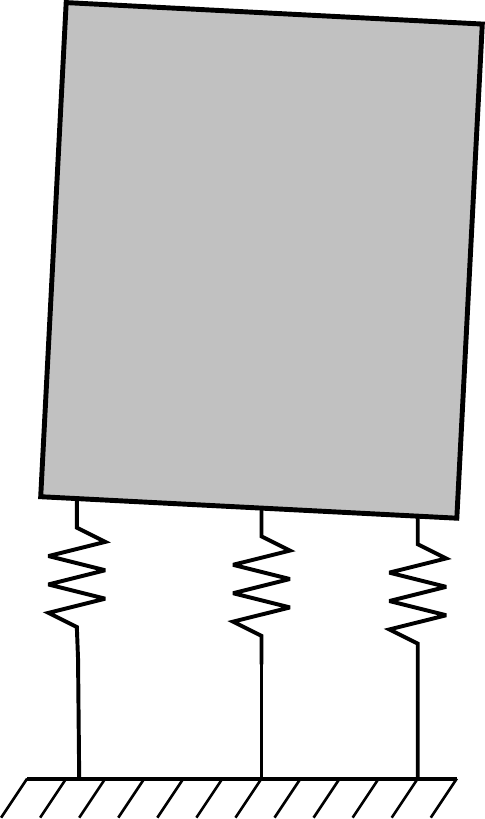}
  }
  
  \caption{Collision detection between a cylinder and a convex-decomposed shape generates one contact point in case 1 ((a), (b) and (c)) and three contact points in case 2 ((d), (e), and (f)) with a slight change in the cylinder's pose. With finer decomposition, the number of contacts may vary significantly. Each contact point can be interpreted as a spring connecting the two bodies to prevent interpenetration. Since each spring adds up to the total stiffness, the system may become overstiff if care is not taken.}
  \label{fig-example-contact}
\end{figure}

In this paper, we present a contact reduction method with bounded stiffness to improve the simulation accuracy. Our method is beneficial when the simulation consists of stiff rigid bodies, in which cases we argue that the number of contact points greatly influences simulation accuracy. Compared to previous works, our method includes an additional post-processing step, which relies on the concept of \emph{contact stiffness}. We show that the proposed method enables training RL policy for a tight-clearance double pin insertion task and successfully deploying the policy on a rigid, position-controlled robot.





\section{RELATED WORKS} \label{sec-related-work}

\textbf{Reinforcement learning for robot assembly} There have been numerous methods proposed for learning robot assembly skills using RL. These works can be categorized into approaches that train RL policy on the physical robot directly \cite{inoueDeepReinforcement2017}, \cite{schoettlerDeepReinforcement2020}, \cite{luoReinforcementLearning2019}, \cite{zhaoOfflineMetaReinforcement2022}, and approaches that follow simulation-to-reality methodology \cite{schoettlerMetaReinforcementLearning2020}, \cite{beltran-hernandezVariableCompliance2020}, \cite{haoMetaResidualPolicy2022}. Most works consider a torque-controlled robot manipulator. Specifically for industrial, position-controlled robots, Hernandez et al. \cite{beltran-hernandezLearningForce2020} raised the safety problem during training and presented a method to learn policy directly on the real robot. We are not aware of any work that performs sim-to-real transfer on a position-controlled robot. 

Most sim-to-real works approximate the actual geometry of the objects with a much simpler one. A cubical peg and a cubical hole is perhaps the most common approximation. Policies trained on this cubical peg-in-hole task are then deployed on other insertion tasks involving mating parts of different shapes (round, triangle, semicircular, hexagon), cables (USB, waterproof connector plug), or gears \cite{schoettlerMetaReinforcementLearning2020}, \cite{beltran-hernandezVariableCompliance2020}, \cite{haoMetaResidualPolicy2022}.

General geometric representations, such as convex decomposition and triangular meshes, provide a fine approximation for complex, concave surfaces. However, this is also associated with the risk of numerical instability and expensive computation. Narang et al. \cite{narangFactoryFast2022} proposed a general framework for the simulation of complex assembly tasks given a representation of the mating parts. They demonstrated successful RL training in simulation for the complex nut and bolt assembly task, but have not yet performed sim-to-real deployment. In \cite{sonSimtoRealTransfer2020}, the contact boundary surface of two parts is directly parameterized with a Neural Network (NN) and learned through supervised learning. The work also demonstrates successful sim-to-real transfer for the nut and bolt assembly task.





\textbf{Contact reduction method} Contact reduction is a technique to reduce the number of contact points generated in a physics simulation. Most existing methods follow the clustering principle, in which similar contacts are grouped together, then one or several representative contacts are chosen/recomputed for each group. Several metrics for contact similarity and clustering algorithms have been proposed. In \cite{hauserRobustContact2016}, the groups are determined through k-means clustering or hierarchical clustering. The distance metric is based on contact normal and contact position. In the same vein, \cite{otaduyModularHaptic2006} also performs k-means clustering, but uses a different distance metric based on contact position. In \cite{narangFactoryFast2022}, contact patches are iteratively created and the remaining contact points are assigned to each patch based on normal similarity. The proposed contact reduction methods improve simulation stability and avoid expensive computation. However, the accuracy of such methods has not been explored.




\section{BACKGROUND} \label{sec-background}

\subsection{Contact simulation pipeline and contact clustering}

A typical rigid body contact simulation pipeline consists of three main steps
\begin{itemize}
  \item Collision detection checks whether two bodies overlap.
  \item Contact generation determines a representation for the contact region, commonly in the form of a finite set of contact points. A contact point is defined by its position, normal direction, and possibly penetration depth. 
  \item Contact response finds motion of rigid bodies to prevent interpenetration. Solution methods include complementarity-based approaches \cite{stewartImplicitTimeStepping1996} and complementarity-free approaches \cite{todorovMuJoCoPhysics2012}.
\end{itemize}

If two bodies are represented by mesh, collision detection algorithms often work by decomposing meshes into primitive shapes (e.g. cylinder, box, sphere), triangles, or convex parts. Contact points are then generated independently for each of these elements. In this way, many contact points can be generated.

The number of contact points greatly influences the accuracy, stability, and speed of the simulation. A reasonable number of contact points are needed to accurately realize the contact region, while too many contact points lead to expensive simulation and may cause numerical instability \cite{otaduyModularHaptic2006}, \cite{narangFactoryFast2022}.

To avoid expensive simulations and improve simulation stability, contact clustering reduces the number of contact points obtained by the contact generation step. This method works by putting similar contacts into groups using some heuristics, then choosing/recomputing from each group one or several representative contact points.

\subsection{Reinforcement learning}

In RL, an agent learns to maximize the total reward received through interacting with its environment. A discounted episodic RL problem can be formalized as a Markov Decision Process (MDP) \cite{suttonReinforcementLearning2018a}. In MDP, an agent interacts with its environment in discrete time steps. At each time step $t$, the agent
observes current state $\bs{s}_t \in \mathcal{S}$, executes an action $\bs{a}_t \in \mathcal{A}$, and receives an immediate reward $r_t$. The environment evolves through the state transition probability $p(\bs{s}_{t+1} | \bs{s}_t, \bs{a}_t)$. The goal in RL is to learn a policy $\pi (\bs{a}_t|\bs{s}_t)$ that maximizes the expected discounted return $R = \sum_{t=1}^{T} \gamma^t r_t$, where $\gamma$ is the discount factor.

\section{CONTACT REDUCTION WITH BOUNDED STIFFNESS} \label{sec-contact-reduction}



Our proposed  contact reduction method consists of two steps. In the first step, k-means clustering is used to reduce the number of contact points to a user-defined number \(k\). In the second step, the contact stiffnesses of all the contact points are determined by a quadratic program, such that the net stiffness is upper bounded.

\subsection{Example: Direct force control of a position-controlled manipulator} \label{sec-example}

To see why the net stiffness should be bounded, consider the direct force control of a position-controlled manipulator. Model-based design methods require a model of the environment. A common model considers the robot as a single point or a small region and represents the environment as an \(n\)-dimensional spring (\(n\leq 6\)), or a spring and a damper \cite{deschutterCompliantRobot1988}, \cite{stoltAdaptationForce2012}, \cite{phamConvexController2020}. In the former case, the model has the following form
\begin{equation}
  \bs{F}_e=\bs{K}_e(\bs{x}_e-\bs{x})
  \label{eq-contact-model-1}
\end{equation}
where \(\bs{K}_e\) is n-dimensional matrix representing the stiffness of the environment, \(\bs{x}_e\) is the robot position just before contact. It is usually assumed that the environment stiffness in different directions are uncoupled. In this case, \(\bs{K}_e\) is a diagonal matrix, and equation (\ref{eq-contact-model-1}) is replaced by \(n\) scalar equations
\begin{equation}
  F_e=K_e(x_e-x)
\end{equation}
Controllers designed using this simple model has proven to be effective even in applications involving complex contact scenarios such as hand guiding \cite{phamConvexController2020}, assembly \cite{stoltAdaptationForce2012}. When the stiffness \(K_e\) is unknown, it can be estimated online or offline \cite{ericksonContactStiffness2003}.

While the above contact model is useful for controller design and analysis, it is too simple for simulation purpose. In this context, the robot-environment interaction is typically represented by a finite number of contact points. Assuming frictionless contact, the contact force at each contact point is
\begin{equation}
  \bs{F}_i=F_{ni}\bs{n}_i
  \label{eq-contact-model-2-1}
\end{equation}
where $\bs{n}_i\in \mathbb{R}^3$ is the contact normal. The contact force between robot and environment can then be computed
\begin{equation}
  \bs{F}_e=\sum{\bs{F}_i}
  \label{eq-contact-model-2-2}
\end{equation}

Many approaches have been proposed for the computation of \(F_n\). In this example, we focus on the spring model which defines the normal component of contact force as follows
\begin{equation}
  F_{ni}=K_i\delta_i
  \label{eq-contact-model-2-3}
\end{equation}
where \(K_i\) is the \emph{contact stiffness}, \(\delta_i\) is the penetration depth calculated by the contact generation step. It's common to use a single value for all contact points.

The penetration depth relates to the robot position \(\bs{x}\) through the equation
\begin{equation}
  \delta_i=\bs{n}_i^T(\bs{x}-\bs{x}_e)
  \label{eq-contact-model-2-4}
\end{equation}
From (\ref{eq-contact-model-2-1})-(\ref{eq-contact-model-2-4}), the contact force can be written as 
\begin{equation}
  \bs{F}_e=K\sum{\bs{n}_i\bs{n}_i^T}(\bs{x}-\bs{x}_e)
  \label{eq-contact-model-2-5}
\end{equation}
which has the form of \ref{eq-contact-model-1} with the equivalent environment stiffness 
\begin{equation}
  \bs{K}_e=K\sum{\bs{n}_i\bs{n}_i^T}  
  \label{eq-contact-model-2-6}
\end{equation}

From equation (\ref{eq-contact-model-2-6}), it can be inferred that the net stiffness may vary significantly during the course of simulation when the contact points change position and normal direction. Specifically, the stiffness along one direction can be any value in the range \([0, NK]\) where \(N\) is the number of contact points. Therefore, the environment may appear stiffer or softer depending on the simulation state. For example, consider the collision between a cylinder and a convex-decomposed object as shown Fig~\ref{fig-example-contact}, convex decomposition may cause "redundant" contact points which add stiffness to the system.

\subsection{Contact clustering}

Each contact point is represented by a 6D vector \([\bs{n}, \bs{x}]\) where \(\bs{n}\) is the contact normal, and \(\bs{x}\) is the contact position. The axis-weighted distance metric \cite{hauserRobustContact2016} is used. Specifically the distance between two points \([n_1, p_1]\) and \([n_2, p_2]\) is computed by
\begin{equation}
  d=||n_2-n_1||_2^2 + c||p_2-p_1||_2^2  
\end{equation}

The centers of cluster are initialized with a deterministic version of k-means++ algorithm \cite{arthurKmeansAdvantages2006}. This algorithm helps avoid suboptimal clustering by spreading out the initial clusters. Although the initialization takes extra time, the main k-means algorithm quickly converge and thus the computation time is actually faster.


\subsection{Scaling contact stiffness}

Our main idea is to limit the net stiffness of the rigid bodies. We propose to scale contact stiffness such that the net stiffness of the system in (\ref{eq-contact-model-2-6}) is upper bounded by \(K_{max}\). The scaling coefficients \(s_i\) are computed by solving the following quadratic program
\begin{equation}
  \begin{aligned}
    \min_{s_i} \quad & \sum{(s_i-1)^2}\\
    s.t. \quad & k\sum{s_i\bs{c}_{ij}} \leq K_{max}, \quad j=1,2,3
  \end{aligned}
\end{equation}
where \(\bs{c}_{i}=diag(\bs{n}_i\bs{n}_i^T)\) is the stiffness induced by contact \(i\) (ignore the coupling stiffness between different directions), \(\bs{c}_{ij}\) is the \(j\) component of \(\bs{c}_i\). The scaling coefficient is optimized such that the change in contact stiffness is minimized. In theory, setting \(K_{max}\) to the contact stiffness \(K\) is a good choice. Using a larger value of \(K_{max}\) can also be beneficial for, as an example, learning robust RL policy.






\section{LEARNING CONTACT-RICH TASKS WITH POSITION-CONTROLLED ROBOTS} \label{sec-learning}

\subsection{Modeling of position-controlled robot} \label{sec-modeling}


In position-controlled robots, the joint torques are commanded by a low-level joint position controller. User access to this controller is typically limited or unavailable \cite{royAdaptiveForce2002}. The work of \cite{phamConvexController2020} assumes a decoupled robot dynamics and the dynamics of each joint is modeled as a first-order linear time-invariant system with time delay. An implicit assumption of this model is that the effect of contact force on the inner joint position controller is negligible, which is a common assumption in the literature \cite{royAdaptiveForce2002}. As a direct consequence, the robot is assumed to be much stiffer than the environment. In other words, a very small position error may result in a huge interaction force. Motivated by this model, we model the joint position controller of the robot with a \emph{computed torque controller}

\begin{equation}
  \tau=M(\ddot{q}_d + D(\dot{q}_d-q) + K(q_d-q)) + \tau_{ext} + C(q, \dot{q}) + g(q)
  \label{eq-controller}
\end{equation}
where \(M\) is the joint inertia matrix, \(q,\dot{q}\) is the joint position, joint velocity respectively, \(q_d, \dot{q}_d, \ddot{q}_d\) is the desired joint position, desired joint velocity, and desired joint acceleration respectively, \(\tau_{ext}\) is the external torque, \(C\) is Coriolis and centrifugal torque, and \(g\) is the gravitational torque. \(K\) and \(D\) are diagonal gain matrices. Compared to \cite{phamConvexController2020}, controller (\ref{eq-controller}) also results in a decoupled dynamics, but the dynamics of each joint is a second-order system instead of first-order.

\subsection{Reinforcement learning framework} \label{sec-rl-framework}



The reinforcement learning framework includes two main components: a parallel position/force controller and an RL policy. The parallel position/force controller comprises a velocity control loop and a force control loop. The velocity control loop is simply a feedforward controller \(u_v=\int{v_d}dt\), while the force control loop is a Proportional-Integral controller \(u_f=k_p(f_d-f)+k_i\int{f_d-f}dt\). The outputs of two control loops are added to obtain the commanded Cartesian pose. Finally, the commanded joint position is obtained through Differential Inverse Kinematics.


The action of RL policy is the desired velocity and desired force, the two inputs of the position/force controller. Note that it is a common practice that RL action are chosen to be the input of a high-level controller \cite{beltran-hernandezLearningForce2020}, \cite{bogdanovicLearningVariable2020}, \cite{martin-martinVariableImpedance2019}. The state is the end-effector pose (with axis angle as the representation of orientation) and the external force acting on the end-effector. End-effector poses can be measured from the joint encoders and the known kinematics model, while the external force can be measured using a six-axis force/torque sensor attached between the robot flange and the end-effector.

\subsection{Reinforcement learning algorithm}

The policy is parameterized by a neural network and trained with the Proximal Policy Optimization (PPO) algorithm \cite{schulmanProximalPolicy2017}. PPO also trains an additional value function, which maps observation to the value of the current observation. The value function is also parameterized by a neural network


Since the value function is only used during training, we use an Asymmetric Actor-Critic \cite{pintoAsymmetricActor2018} approach. Asymmetric Actor-Critic exploits the fact that the value function can have access to information that is not available on the real robot system (for instance, error-free poses of objects in the scene). The additional input potentially accelerates the learning of good value estimates since less information needs to be inferred.

\section{EXPERIMENT} \label{sec-exp}

The aims of the experiments are to (1) validate the performance and advantages of the proposed contact reduction method (2) demonstrate sim-to-real transfer for a tight-clearance cylindrical pin insertion task and a double pin insertion task.





\subsection{Contact reduction performance} \label{sec-contact-reduction-performance}

\begin{figure}[!t]
  \centering
  \subfloat[]{
    \includegraphics[width=0.28\columnwidth]{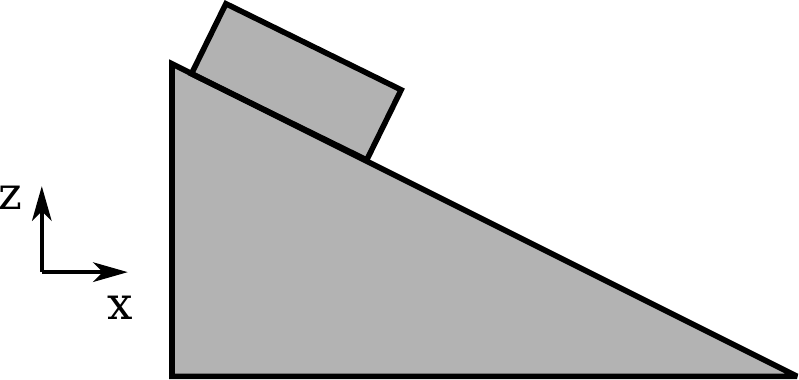}
    \label{fig-example-accuracy-a} }\,
  \subfloat[]{
    \includegraphics[width=0.28\columnwidth]{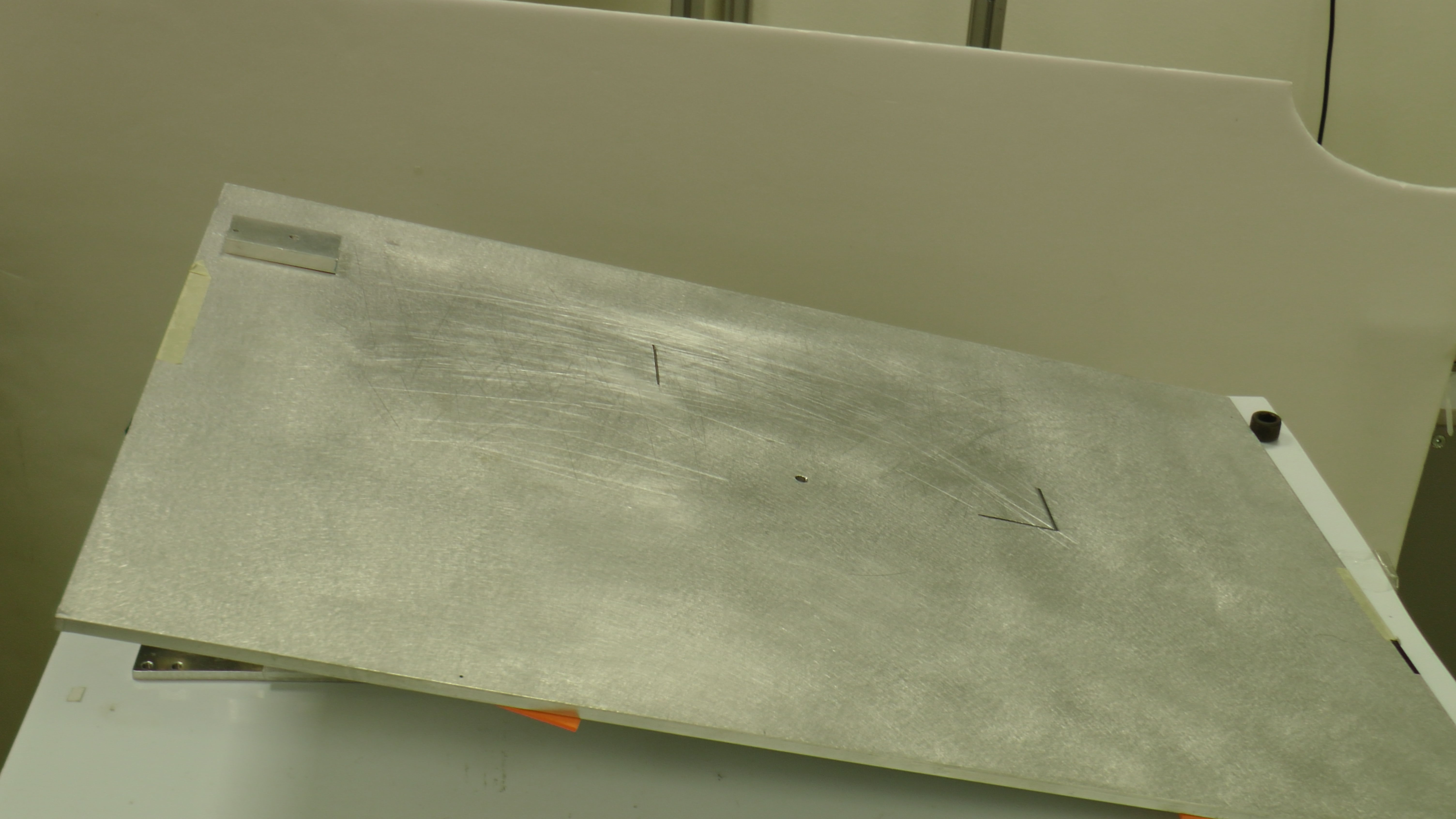}
    \label{fig-example-accuracy-d} }\,
  \subfloat[]{
    \includegraphics[width=0.28\columnwidth]{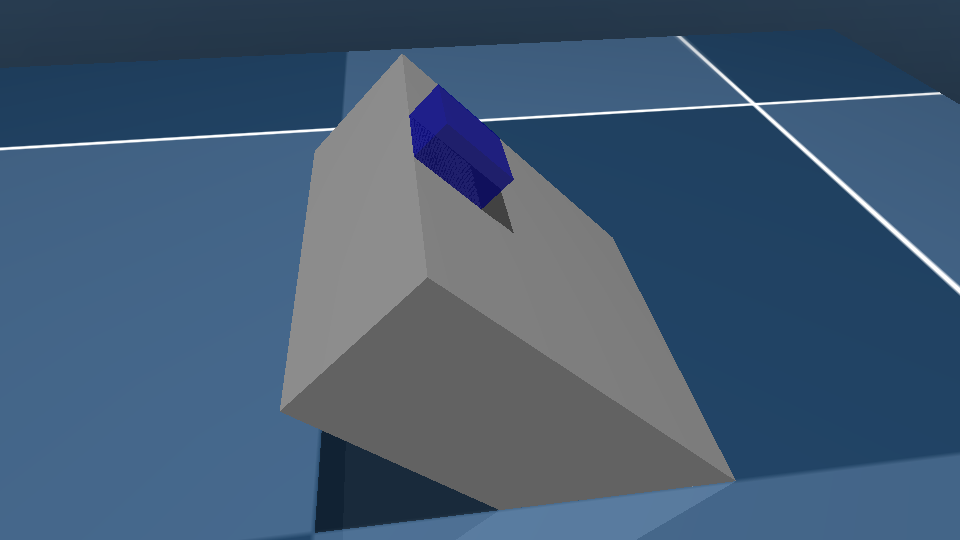}
    \label{fig-example-accuracy-b} }\\
  \subfloat[]{
    \includegraphics[width=0.95\columnwidth]{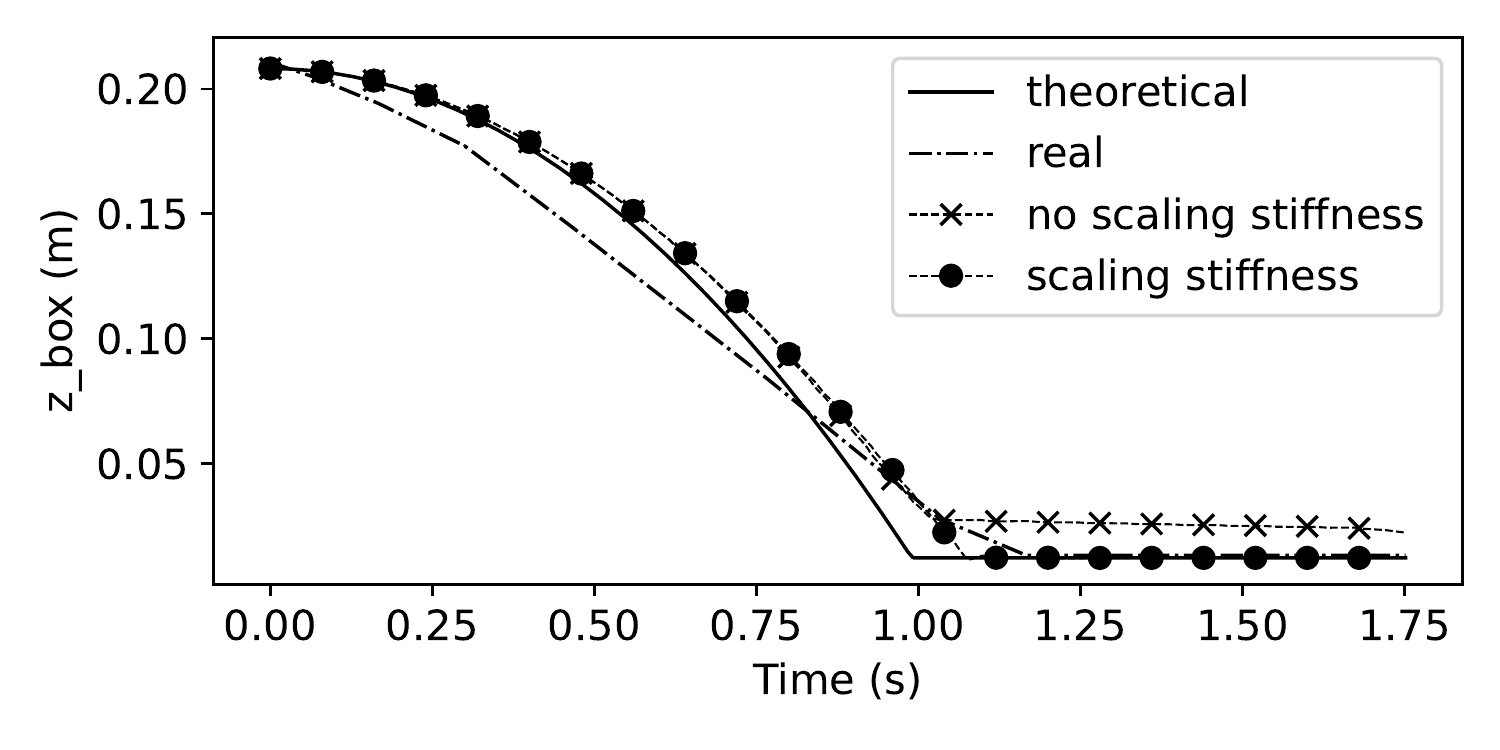}
    \label{fig-example-accuracy-c} }

  \caption{(a) Illustration of a simple example: a box slides down an inclined plane, (b) Real experiment, (c) The example is simulated in Mujoco, (d) The evolution of the position of the box center along the z axis over time in three cases. With scaling stiffness, the trajectory of the box position closely matches theoretical solution. Without scaling stiffness, the box stuck on the plane.}
  \label{fig-example-accuracy}
\end{figure}


\begin{table*}[]
  \centering
  \begin{tabular}{c|c|cc|cc}
  \multirow{2}{*}{}     & \multirow{2}{*}{\begin{tabular}[c]{@{}c@{}}Number of \\ contact points\end{tabular}} & \multicolumn{2}{c|}{Collision detection time (ms)}     & \multicolumn{2}{c}{Contact response time (ms)}         \\ \cline{3-6} 
                        &                                                                                      & \multicolumn{1}{c|}{Baseline} & Proposed & \multicolumn{1}{c|}{Baseline} & Proposed \\ \hline
  USB insertion         & 16                                                                                   & \multicolumn{1}{c|}{1.071}    & 1.072                  & \multicolumn{1}{c|}{0.74}     & 0.22                   \\ \hline
  Round pin insertion   & 21                                                                                   & \multicolumn{1}{c|}{0.23}     & 0.232                  & \multicolumn{1}{c|}{0.17}     & 0.034                  \\ \hline
  Nut and bolt assembly & 204                                                                                  & \multicolumn{1}{c|}{16.4}     & 16.402                 & \multicolumn{1}{c|}{0.45}     & 0.24                  
\end{tabular}
  \caption{Influence of the the proposed method on simulation speed}
  \label{tab-contact-reduction-performance}

\end{table*}

The proposed contact reduction method was implemented into Mujoco physics engine \cite{todorovMuJoCoPhysics2012} and evaluated in several scenarios. First we show that the proposed method improve the accuracy of multiple-contacts simulation in a simple, yet ubiquitous case. The scenario includes a box sliding on a inclined plane as shown in Fig~\ref{fig-example-accuracy-a}. In Mujoco, this scenario can be simulated using the box shapes (see Fig~\ref{fig-example-accuracy-b}). Using this primitive shape generates at most four contact point when the box slide on the surface. To simulate multiple contacts, the inclined plane was modeled with a mesh instead of the primitive box shape. The mesh was decomposed into smaller parts, each of which can generate one contact point with the box. The decomposition was done in such a way that the number of contact points increase as the box slides down the plane and at most 512 contact points could be generated at a time. Note that the decomposition was done for the purpose of evaluating the method in multiple-contacts simulation. Although the decomposition is unnecessary to model a plane, it is required to model objects containing concave features such as holes.

We run simulation in two settings: without scaling stiffness, and with scaling stiffness. The evolution of the box center's position is recorded in Fig~\ref{fig-example-accuracy-c}. The theoretical solution can be easily obtained by Newton's second law and is used as the reference. We also carried out a real experiment for this problem. The progress was recorded by a Panasonic HC-X920M camera. The box position were then estimated roughly from the video. With scaling stiffness, the trajectory of the box center closely follow theoretical solution. The box eventually reaches the ground, which comply with the theoretical and real results. On the other hand, the box stuck on the plane without scaling stiffness. The reason is that the contact force between the cylinder and the plane became too large, causing a large friction force. On the other hand, our method maintains the contact force regardless of the number of contacts, thus make the simulator more realistic.

Next we evaluate how the proposed method affect simulation speed in several assembly scenes. Each scene includes a 6-axis Denso VS-060 robot and two mating parts. The joint position controller and the hybrid motion/force controller were implemented as described in Section~\ref{sec-modeling} and Section~\ref{sec-rl-framework}, respectively. The collision geometries of the mating parts were modeled with primitive shapes (cylinders, boxes, spheres) if possible; otherwise, they were decomposed into multiple convex hulls manually or by V-HACD \cite{lengyelVolumetricHierarchical2016}. One part was attached rigidly to the robot end-effector (grasping was not simulated), and the other part was rigidly placed in the environment. The scenes are described as follows

\begin{figure}[!t]
  \centering
  \subfloat[USB insertion]{
    \includegraphics[width=0.3\columnwidth]{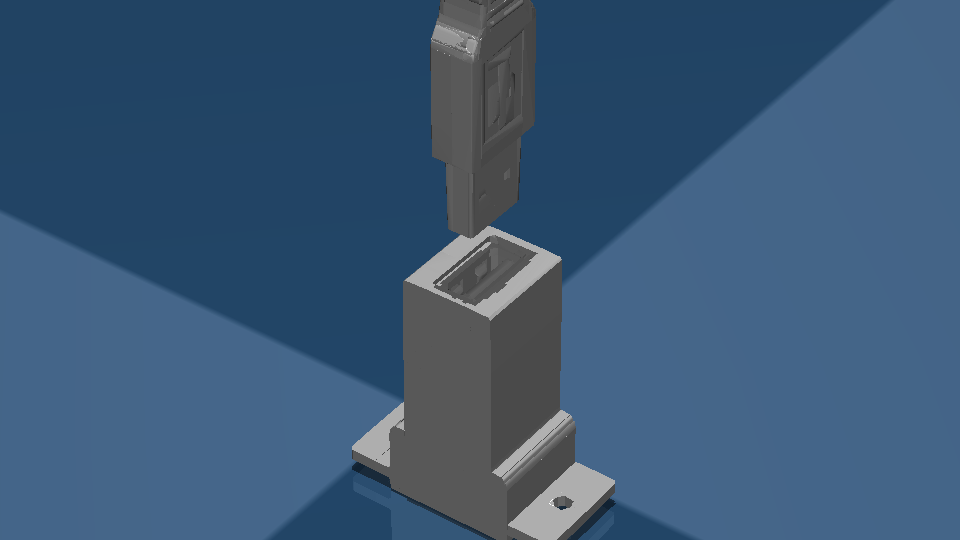}
    \label{fig-example-speed-a} }
  \subfloat[Pin insertion]{
    \includegraphics[width=0.3\columnwidth]{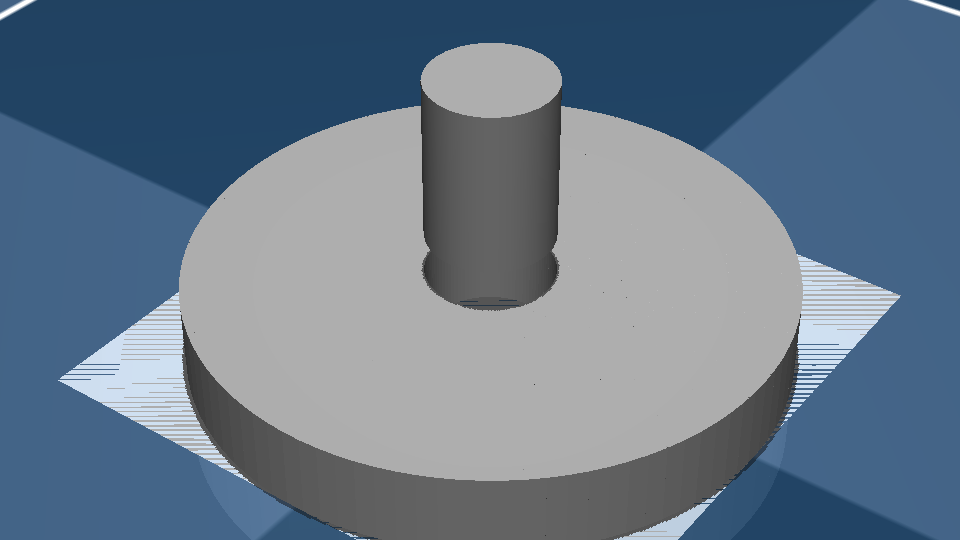}
    \label{fig-example-speed-b} }
  \subfloat[Nut and bolt assembly]{
    \includegraphics[width=0.3\columnwidth]{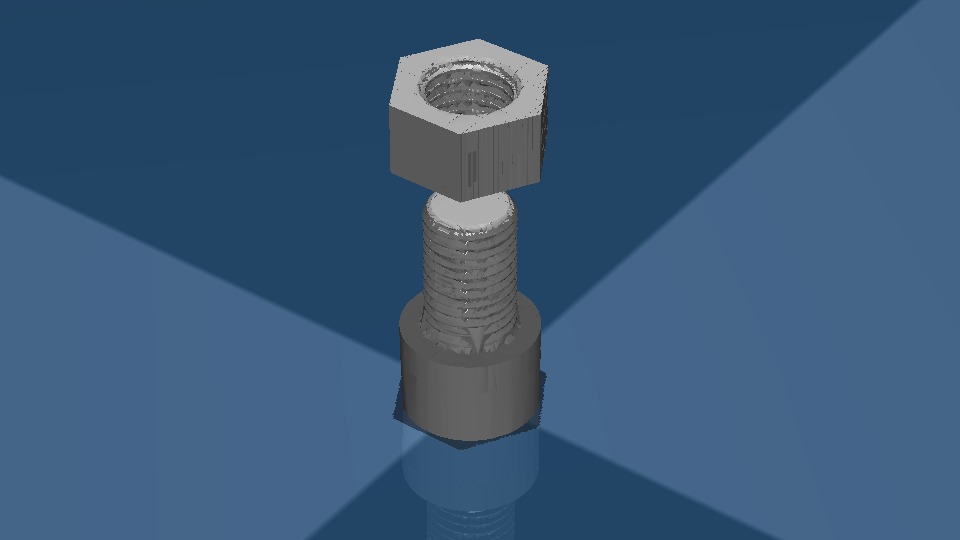}
    \label{fig-example-speed-c} }\\

  \caption{Three assembly scenes in Mujoco}
  \label{fig-example-speed}
\end{figure}

\textbf{USB insertion} (Fig~\ref{fig-example-speed-a}) The USB female and male meshes were sourced from the manufacturer. Both parts were decomposed into 200 pieces using V-HACD.


\textbf{Round peg insertion} (Fig~\ref{fig-example-speed-b}) The round peg was modeled using the cylinder shape in Mujoco. The round hole was manually decomposed into 100 convex hulls.

\textbf{Nut and bolt assembly} (Fig~\ref{fig-example-speed-c}) Both the nut and bolt were decomposed into 1500 pieces using V-HACD


The number of clusters was set to 10 in all scenes. We manually designed motion scripts to simulate the insertion phase and record the average number of contacts, average collision detection time, and average contact response time over all simulation steps in Table~\ref{tab-contact-reduction-performance}. We observe that the proposed contact reduction algorithm adds a very small extra time (about 0.002\,ms in the three scenes) to the collision detection time. Owing to the reduced number of contacts, our method reduces the contact response time by 0.52\,ms, 0.136\,ms, and 0.21\,ms in the USB insertion, round peg insertion, and the nut and bolt assembly, respectively. Since the gain in contact response performance far exceeds the loss in the collision detection, the proposed method significantly improve simulation speed.




\subsection{Reinforcement learning and sim-to-real tranfer result} \label{sec-sim2real-result}


In this section, we show successful sim-to-real transfer results for two tasks: the round pin insertion and the double pin insertion task. The round pin insertion task is similar to the scene used in the previous section. The double pin insertion task is more challenging due to the yaw error. The number of clusters was set to four for the round pin insertion task and ten for the double pin insertion tasks. The maximum net stiffness \(K_{max}\) was set to twofold the unscaled contact stiffness.

\begin{figure}[!t]
  \centering
  \includegraphics[width=\columnwidth]{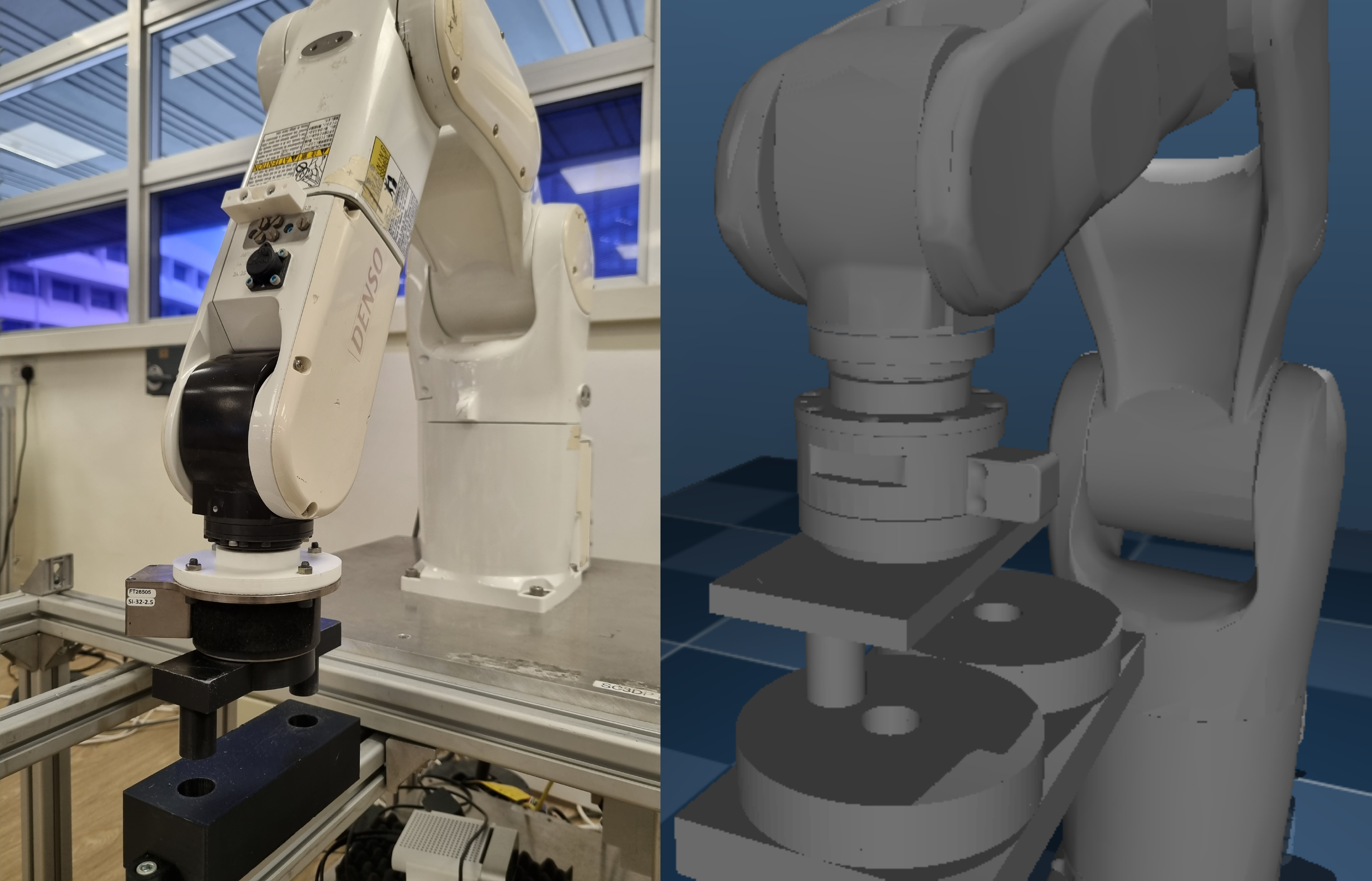}
  \caption{Hardware setup (left) and the corresponding simulated environment of the double pin insertion task (right)}
  \label{fig-setup}
\end{figure}

Real experiments were carried out on a 6-axis Denso VS-060 robot. An ATI Gamma force/torque sensor was rigidly attached to the end-effector to provide contact force measurements. A personal computer running Ubuntu 16.04 was used to send commands to the robot and train RL policy. For both tasks, the pegs were rigidly attached to the force torque sensor. The holes were rigidly mounted within the robot's workspace. For the round pin insertion task, the peg and hole were made of aluminum with 0.05\,mm clearance. For the double pin insertion, the parts were 3D printed and the clearance was measured to be 0.2\,mm. The hardware setup and corresponding simulated environment for the double pin insertion task are shown in Fig~\ref{fig-setup}

For both tasks, the following reward function was used
\begin{equation}
  r=\frac{z-z_0}{D}-1+r_f
\end{equation}
\begin{equation}
r_f=\begin{cases}
  -2 &\text{if } \max(\bs{f} - \bs{f}_{u})>0\\
  0 &\text{otherwise}
\end{cases}  
\end{equation}
where \(z\) is the end-effector position along the \(z-\)axis (assumed to be the insertion axis), \(z_0\) is the end-effector position when it first touches the hole (i.e. when a force along the \(z-\)axis is sensed), \(D\) is the insertion depth, \(\bs{f}\) and \(\bs{f}_u\) is the measured contact force and the upper limit force. 



Each policy was trained for each task for 500 epochs. At the beginning of each training episode, the robot is reset to a random configuration such that the relative position between the peg and the hole is less than 2\,mm (except for the insertion direction) and the relative orientation error is less than 1\,deg. 

After training, each policy was evaluated 20 times on the real robot. The success rate and average completion time are reported in Table~\ref{tab-sim2real-result}. The only failure in the double pin insertion task is because the peg drifts away from the initial pose and fails to recover. The learned strategy is also quite intuitive. The peg is tilted to easier align with the hole rim, followed by an oscillating motion to insert the peg.

\begin{table}[!t]
  \caption{Sim-to-real result}
  \begin{tabular}{l||c|c}
  Task & Success rate & Average completion time (s) \\ \hline
  Round pin insertion & 1 & 2.31 \(\pm\) 0.47  \\
  Double pin insertion & 0.95 & 2.87 \(\pm\) 0.33
  \end{tabular}
  \label{tab-sim2real-result}
\end{table}


\subsection{Effect of scaling contact stiffness} \label{sec-effect-scaling}

First, we experimentally show that the environment may appear stiffer due to the increase in the number of contacts as described in Section~\ref{sec-example}. The round pin insertion scene was used for this purpose. The robot was reset to a configuration such that the angle between the peg's surface and the hole's surface is one degree, then commanded to come into contact with the hole surface along the surface's normal. A constant desired force of 5\,N and 30\,N is successively sent to the force controller. The number of contact points depends on the number of convex shapes composing the hole.



\begin{figure}[!t]
  \centering
  \subfloat[Four contact points]{
    \includegraphics[width=0.45\columnwidth]{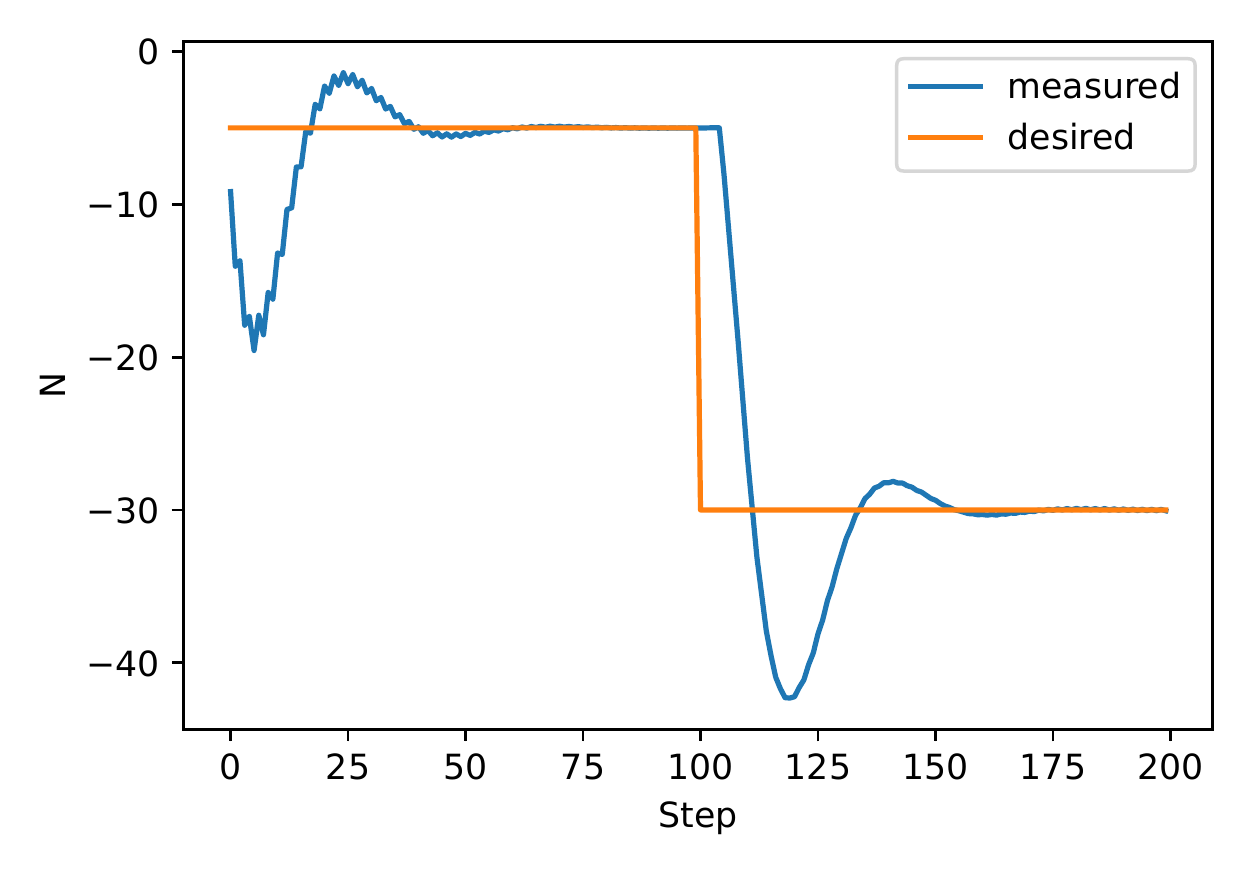}
  }
  \subfloat[Six contact points]{
    \includegraphics[width=0.45\columnwidth]{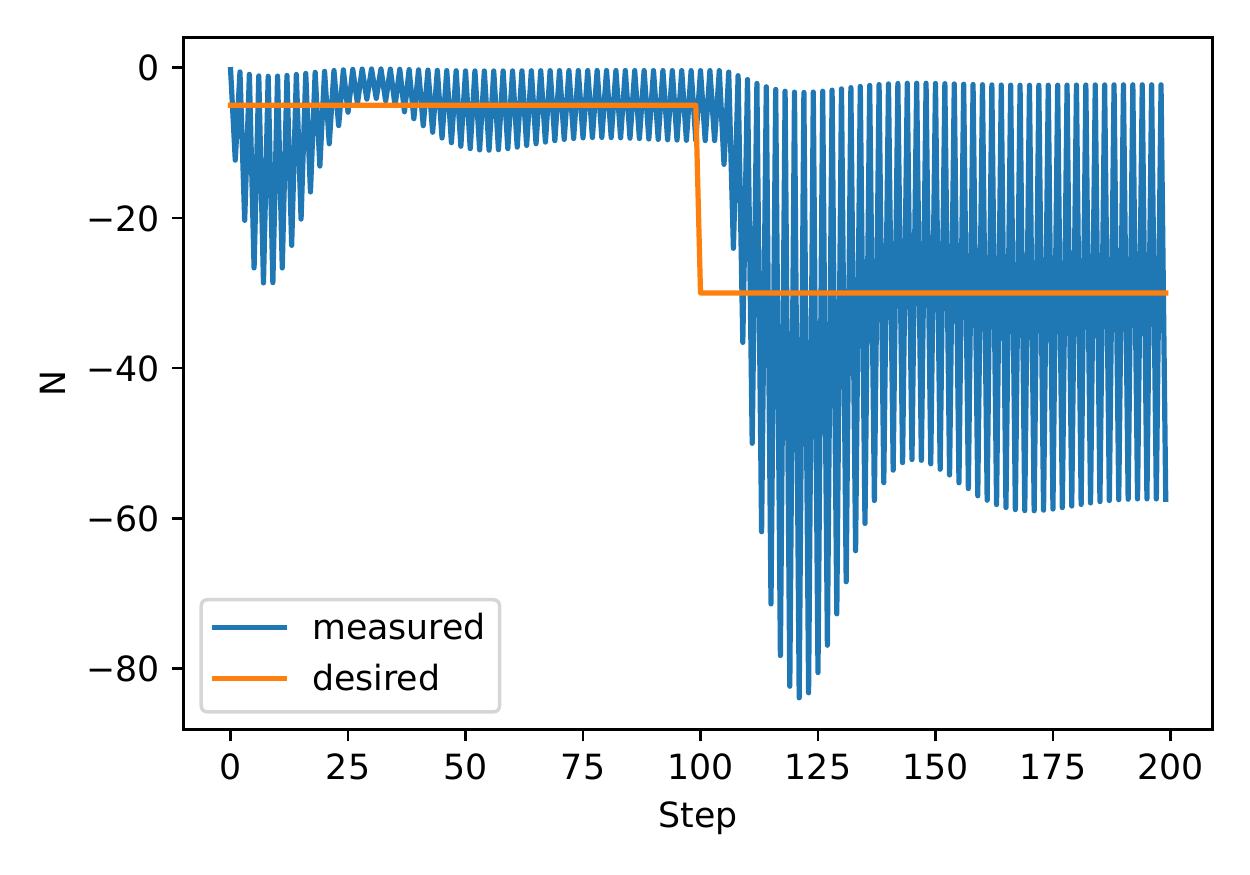}
  }
  \caption{Without scaling stiffness, the force controller becomes instable when the number of contact increases. The robot is commanded such that the peg comes into contact with the hole surface along the surface's normal. Input desired forces of 5\,N and 30\,N are then sequentially sent to the force controller.}
  \label{fig-force-response}
\end{figure}

As shown if Fig~\ref{fig-force-response}, the force controller becomes instable with six contact points. To prevent instability, the number of clusters should be four or less. However, this may compromise simulation accuracy. For instance, consider the double pin insertion task, at least six contact points are required to account for all possible contact configurations.

\begin{figure}[!t]
  \centering
  \subfloat[Success rate]{\includegraphics[width=\columnwidth]{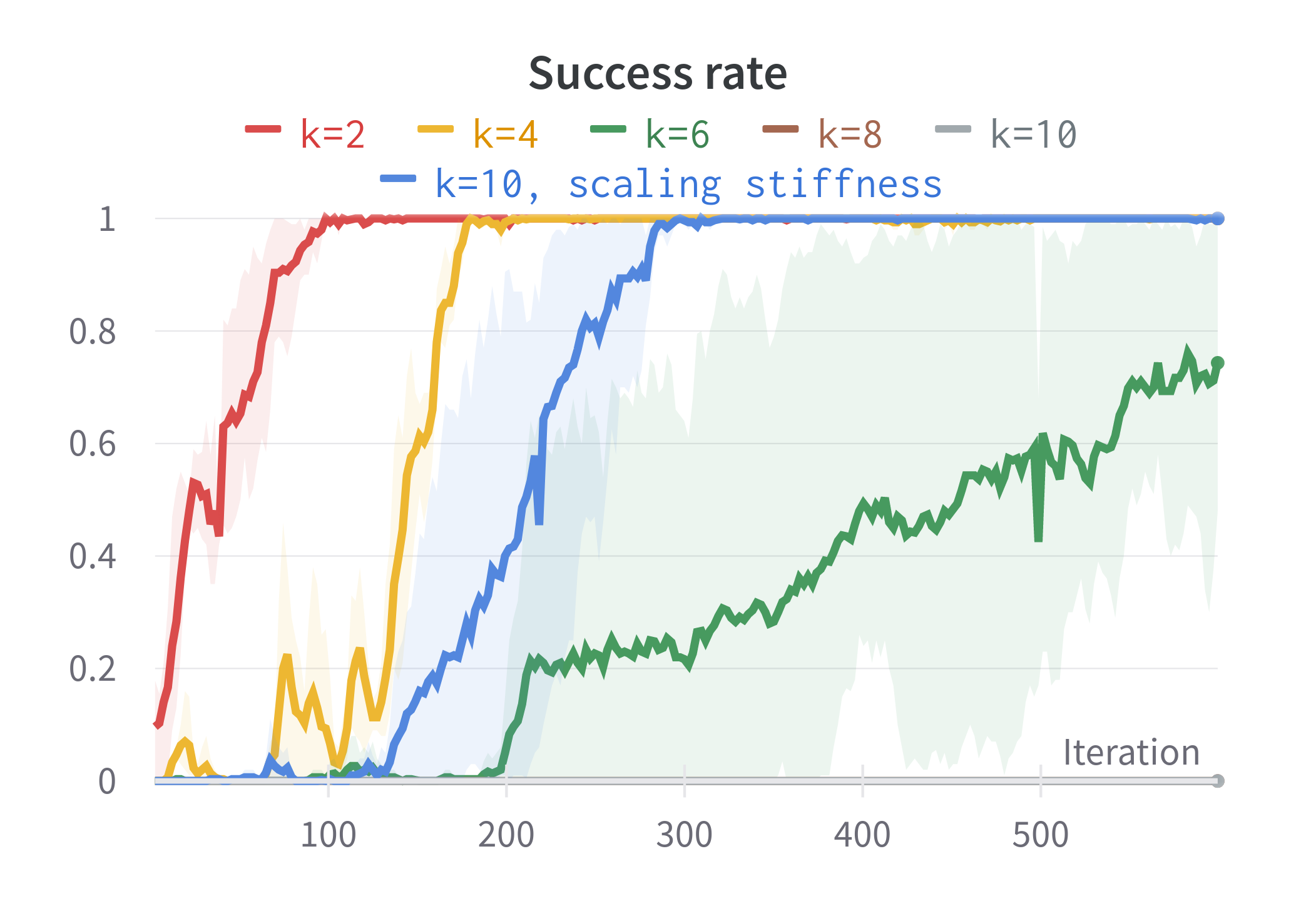}}\\
  \subfloat[Return]{\includegraphics[width=\columnwidth]{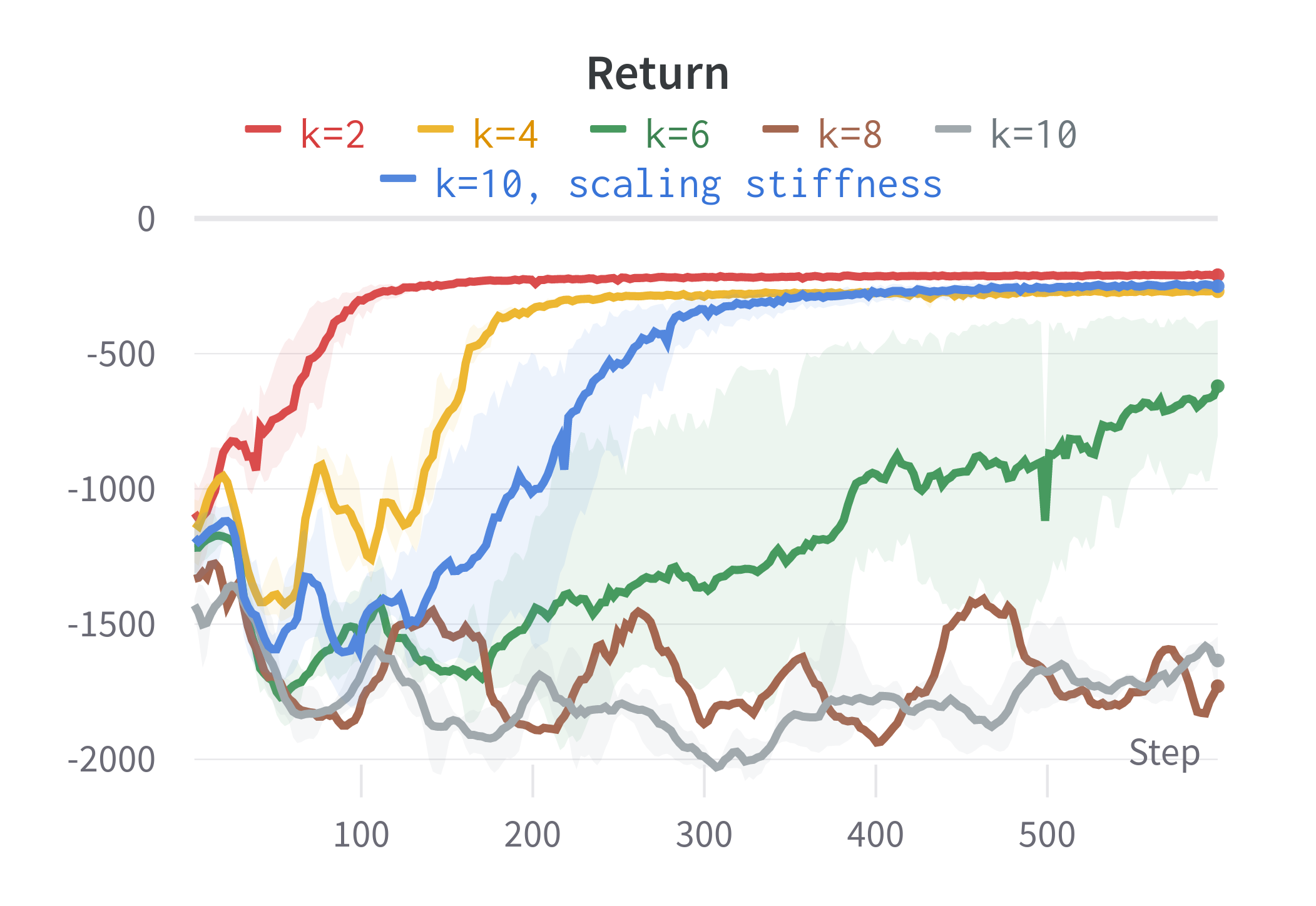}}
  \caption{Comparison of training performance for the round pin insertion task for different number of clusters \(k\) in the set \({2,4,6,8,10}\) without scaling stiffness. We also show training performance for \(k=10\) with scaling stiffness (the proposed method).
  }
  \label{fig-training-curve}
\end{figure}

We also explore how the number of contact points influences RL training performance. Several policies were trained for different numbers of clusters in the set \({2,4,6,8,10}\) without scaling stiffness. The training curve is shown in Fig~\ref{fig-training-curve}. As the number of clusters increases, RL training converges slower and the final success rate and return also decline. With \(k\geq 6\) clusters, the final success rate can't reach 100\,\%. The worse RL performance is possibly due to simulation instability, causing the force/torque reading to be chaotic.

Three policies reaching 100\,\% success rate are tested on the real robot with the same procedure in Section~\ref{sec-sim2real-result}. The results are reported in Table~\ref{tab-sim2real-compare}

\begin{table}[!t]
  \caption{Sim-to-real result for the double pin insertion task}
  \begin{tabular}{l||c|c}
   & Success rate & Average completion time \\ \hline
  2 clusters & 0 & - \\
  4 clusters & 1 & 3.63 \(\pm\) 0.72 \\
  10 clusters, scaling stiffness & 0.95 &  2.87 \(\pm\) 0.33 
  \end{tabular}
  \label{tab-sim2real-compare}
\end{table}

The policy trained with \(k=2\) clusters can only partially insert the peg but is unable to fully insert it. A possible reason is that during the insertion phase, the most dominant contact configuration is four-point contact, so two clusters are not enough to accurately represent the actual contact configuration. This manifests into inaccurate contact force/torque reading, causing the policy to fail to transfer.

Suprisingly, the policy trained with \(k=4\) clusters achieves 100\% success rate. This may be because the five- or six-point contact configurations only happen when the peg is tilted, which rarely occurs. As the six-point contacts are indistinguishable from four-point contacts, the learned policy mostly selects translational motion during the search phase, in contrast to the tilting strategy learned by the policy trained with \(k=10\) clusters. The translational motion during the search phase seems to be random, which explains the worse completion time than the tilting strategy.


\section{CONCLUSIONS} \label{sec-conclusion}

In this paper, we have presented a contact reduction method with bounded stiffness to improve the simulation accuracy. Our method is particularly beneficial when simulating interaction between stiff objects. Our experimental results have shown that the proposed method improves simulation speed and that it enables training RL policy in simulation and deploying the trained policy on a challenging double pin insertion task using a position-controlled robot. In future works, we plan to extensively evaluate this method in different simulation scenarios and apply this method for learning different assembly tasks, such as cable insertion.



\addtolength{\textheight}{-12cm}







\section*{ACKNOWLEDGMENT}

This research was supported by the National Research Foundation, Prime
Minister’s Office, Singapore under its Medium Sized Centre funding
scheme, Singapore Centre for 3D Printing, CES\_SDC Pte Ltd, and Chip
Eng Seng Corporation Ltd.


\bibliographystyle{IEEEtran}
\bibliography{ref}

\begin{thebibliography}{10}
\providecommand{\url}[1]{#1}
\csname url@samestyle\endcsname
\providecommand{\newblock}{\relax}
\providecommand{\bibinfo}[2]{#2}
\providecommand{\BIBentrySTDinterwordspacing}{\spaceskip=0pt\relax}
\providecommand{\BIBentryALTinterwordstretchfactor}{4}
\providecommand{\BIBentryALTinterwordspacing}{\spaceskip=\fontdimen2\font plus
\BIBentryALTinterwordstretchfactor\fontdimen3\font minus
  \fontdimen4\font\relax}
\providecommand{\BIBforeignlanguage}[2]{{%
\expandafter\ifx\csname l@#1\endcsname\relax
\typeout{** WARNING: IEEEtran.bst: No hyphenation pattern has been}%
\typeout{** loaded for the language `#1'. Using the pattern for}%
\typeout{** the default language instead.}%
\else
\language=\csname l@#1\endcsname
\fi
#2}}
\providecommand{\BIBdecl}{\relax}
\BIBdecl

\bibitem{andrychowiczLearningDexterous2020}
\BIBentryALTinterwordspacing
O.~M. Andrychowicz, B.~Baker, M.~Chociej, R.~Józefowicz, B.~McGrew,
  J.~Pachocki, A.~Petron, M.~Plappert, G.~Powell, A.~Ray, J.~Schneider,
  S.~Sidor, J.~Tobin, P.~Welinder, L.~Weng, and W.~Zaremba, ``Learning
  dexterous in-hand manipulation,'' vol.~39, no.~1, pp. 3--20. [Online].
  Available: \url{http://journals.sagepub.com/doi/10.1177/0278364919887447}
\BIBentrySTDinterwordspacing

\bibitem{hauserRobustContact2016}
\BIBentryALTinterwordspacing
K.~Hauser, ``Robust {{Contact Generation}} for {{Robot Simulation}} with
  {{Unstructured Meshes}},'' in \emph{Robotics {{Research}}: {{The}} 16th
  {{International Symposium ISRR}}}, ser. Springer {{Tracts}} in {{Advanced
  Robotics}}, M.~Inaba and P.~Corke, Eds.\hskip 1em plus 0.5em minus
  0.4em\relax {Springer International Publishing}, pp. 357--373. [Online].
  Available: \url{https://doi.org/10.1007/978-3-319-28872-7_21}
\BIBentrySTDinterwordspacing

\bibitem{narangFactoryFast2022}
Y.~Narang, K.~Storey, I.~Akinola, M.~Macklin, P.~Reist, L.~Wawrzyniak, Y.~Guo,
  A.~Moravanszky, G.~State, M.~Lu, A.~Handa, and D.~Fox, ``Factory: {{Fast
  Contact}} for {{Robotic Assembly}},'' in \emph{Proceedings of {{Robotics}}:
  {{Science}} and {{Systems}}}.

\bibitem{otaduyModularHaptic2006}
M.~Otaduy and M.~Lin, ``A modular haptic rendering algorithm for stable and
  transparent 6-{{DOF}} manipulation,'' vol.~22, no.~4, pp. 751--762.

\bibitem{inoueDeepReinforcement2017}
\BIBentryALTinterwordspacing
T.~Inoue, G.~De~Magistris, A.~Munawar, T.~Yokoya, and R.~Tachibana, ``Deep
  reinforcement learning for high precision assembly tasks,'' in \emph{2017
  {{IEEE}}/{{RSJ International Conference}} on {{Intelligent Robots}} and
  {{Systems}} ({{IROS}})}.\hskip 1em plus 0.5em minus 0.4em\relax {IEEE}, pp.
  819--825. [Online]. Available:
  \url{http://ieeexplore.ieee.org/document/8202244/}
\BIBentrySTDinterwordspacing

\bibitem{schoettlerDeepReinforcement2020}
G.~Schoettler, A.~Nair, J.~Luo, S.~Bahl, J.~Aparicio~Ojea, E.~Solowjow, and
  S.~Levine, ``Deep {{Reinforcement Learning}} for {{Industrial Insertion
  Tasks}} with {{Visual Inputs}} and {{Natural Rewards}},'' in \emph{2020
  {{IEEE}}/{{RSJ International Conference}} on {{Intelligent Robots}} and
  {{Systems}} ({{IROS}})}, pp. 5548--5555.

\bibitem{luoReinforcementLearning2019}
J.~Luo, E.~Solowjow, C.~Wen, J.~A. Ojea, A.~M. Agogino, A.~Tamar, and
  P.~Abbeel, ``Reinforcement {{Learning}} on {{Variable Impedance Controller}}
  for {{High-Precision Robotic Assembly}},'' in \emph{2019 {{International
  Conference}} on {{Robotics}} and {{Automation}} ({{ICRA}})}, pp. 3080--3087.

\bibitem{zhaoOfflineMetaReinforcement2022}
T.~Z. Zhao, J.~Luo, O.~Sushkov, R.~Pevceviciute, N.~Heess, J.~Scholz,
  S.~Schaal, and S.~Levine, ``Offline {{Meta-Reinforcement Learning}} for
  {{Industrial Insertion}},'' in \emph{2022 {{International Conference}} on
  {{Robotics}} and {{Automation}} ({{ICRA}})}, pp. 6386--6393.

\bibitem{schoettlerMetaReinforcementLearning2020}
G.~Schoettler, A.~Nair, J.~A. Ojea, S.~Levine, and E.~Solowjow,
  ``Meta-{{Reinforcement Learning}} for {{Robotic Industrial Insertion
  Tasks}},'' in \emph{2020 {{IEEE}}/{{RSJ International Conference}} on
  {{Intelligent Robots}} and {{Systems}} ({{IROS}})}, pp. 9728--9735.

\bibitem{beltran-hernandezVariableCompliance2020}
\BIBentryALTinterwordspacing
C.~C. Beltran-Hernandez, D.~Petit, I.~G. Ramirez-Alpizar, and K.~Harada,
  ``Variable {{Compliance Control}} for {{Robotic Peg-in-Hole Assembly}}: {{A
  Deep-Reinforcement-Learning Approach}},'' vol.~10, no.~19, p. 6923. [Online].
  Available: \url{https://www.mdpi.com/2076-3417/10/19/6923}
\BIBentrySTDinterwordspacing

\bibitem{haoMetaResidualPolicy2022}
P.~Hao, T.~Lu, S.~Cui, J.~Wei, Y.~Cai, and S.~Wang, ``Meta-{{Residual Policy
  Learning}}: {{Zero-Trial Robot Skill Adaptation}} via {{Knowledge Fusion}},''
  vol.~7, no.~2, pp. 3656--3663.

\bibitem{beltran-hernandezLearningForce2020}
\BIBentryALTinterwordspacing
C.~C. Beltran-Hernandez, D.~Petit, I.~G. Ramirez-Alpizar, T.~Nishi, S.~Kikuchi,
  T.~Matsubara, and K.~Harada, ``Learning {{Force Control}} for {{Contact-rich
  Manipulation Tasks}} with {{Rigid Position-controlled Robots}},'' vol.~5,
  no.~4, pp. 5709--5716. [Online]. Available:
  \url{http://arxiv.org/abs/2003.00628}
\BIBentrySTDinterwordspacing

\bibitem{sonSimtoRealTransfer2020}
D.~Son, H.~Yang, and D.~Lee, ``Sim-to-{{Real Transfer}} of {{Bolting Tasks}}
  with {{Tight Tolerance}},'' in \emph{2020 {{IEEE}}/{{RSJ International
  Conference}} on {{Intelligent Robots}} and {{Systems}} ({{IROS}})}, pp.
  9056--9063.

\bibitem{stewartImplicitTimeStepping1996}
D.~Stewart and J.~C. Trinkle, ``An {{Implicit Time-Stepping Scheme}} for
  {{Rigid Body Dynamics}} with {{Coulomb Friction}},'' vol.~39, pp. 2673--2691.

\bibitem{todorovMuJoCoPhysics2012}
E.~Todorov, T.~Erez, and Y.~Tassa, ``{{MuJoCo}}: {{A}} physics engine for
  model-based control,'' in \emph{2012 {{IEEE}}/{{RSJ International
  Conference}} on {{Intelligent Robots}} and {{Systems}}}, pp. 5026--5033.

\bibitem{suttonReinforcementLearning2018a}
R.~S. Sutton and A.~G. Barto, \emph{Reinforcement {{Learning}}, Second Edition:
  {{An Introduction}}}.\hskip 1em plus 0.5em minus 0.4em\relax {MIT Press}.

\bibitem{deschutterCompliantRobot1988}
\BIBentryALTinterwordspacing
J.~De~Schutter and H.~Van~Brussel, ``Compliant {{Robot Motion II}}. {{A Control
  Approach Based}} on {{External Control Loops}},'' vol.~7, no.~4, pp. 18--33.
  [Online]. Available: \url{https://doi.org/10.1177/027836498800700402}
\BIBentrySTDinterwordspacing

\bibitem{stoltAdaptationForce2012}
\BIBentryALTinterwordspacing
A.~Stolt, M.~Linderoth, A.~Robertsson, and R.~Johansson, ``Adaptation of
  {{Force Control Parameters}} in {{Robotic Assembly}},'' vol.~45, no.~22, pp.
  561--566. [Online]. Available:
  \url{https://www.sciencedirect.com/science/article/pii/S1474667016336692}
\BIBentrySTDinterwordspacing

\bibitem{phamConvexController2020}
\BIBentryALTinterwordspacing
H.~Pham and Q.-C. Pham, ``Convex {{Controller Synthesis}} for {{Robot
  Contact}}.'' [Online]. Available: \url{http://arxiv.org/abs/1909.04313}
\BIBentrySTDinterwordspacing

\bibitem{ericksonContactStiffness2003}
\BIBentryALTinterwordspacing
D.~Erickson, M.~Weber, and I.~Sharf, ``Contact {{Stiffness}} and {{Damping
  Estimation}} for {{Robotic Systems}},'' vol.~22, no.~1, pp. 41--57. [Online].
  Available: \url{https://doi.org/10.1177/0278364903022001004}
\BIBentrySTDinterwordspacing

\bibitem{arthurKmeansAdvantages2006}
\BIBentryALTinterwordspacing
D.~Arthur and S.~Vassilvitskii. K-means++: {{The Advantages}} of {{Careful
  Seeding}}. [Online]. Available:
  \url{http://ilpubs.stanford.edu:8090/778/?ref=https://githubhelp.com}
\BIBentrySTDinterwordspacing

\bibitem{royAdaptiveForce2002}
J.~Roy and L.~Whitcomb, ``Adaptive force control of position/velocity
  controlled robots: Theory and experiment,'' vol.~18, no.~2, pp. 121--137.

\bibitem{bogdanovicLearningVariable2020}
\BIBentryALTinterwordspacing
M.~Bogdanovic, M.~Khadiv, and L.~Righetti, ``Learning {{Variable Impedance
  Control}} for {{Contact Sensitive Tasks}}.'' [Online]. Available:
  \url{http://arxiv.org/abs/1907.07500}
\BIBentrySTDinterwordspacing

\bibitem{martin-martinVariableImpedance2019}
R.~Martín-Martín, M.~A. Lee, R.~Gardner, S.~Savarese, J.~Bohg, and A.~Garg,
  ``Variable impedance control in end-effector space: {{An}} action space for
  reinforcement learning in contact-rich tasks.''

\bibitem{schulmanProximalPolicy2017}
\BIBentryALTinterwordspacing
J.~Schulman, F.~Wolski, P.~Dhariwal, A.~Radford, and O.~Klimov, ``Proximal
  {{Policy Optimization Algorithms}}.'' [Online]. Available:
  \url{http://arxiv.org/abs/1707.06347}
\BIBentrySTDinterwordspacing

\bibitem{pintoAsymmetricActor2018}
\BIBentryALTinterwordspacing
L.~Pinto, M.~Andrychowicz, P.~Welinder, W.~Zaremba, and P.~Abbeel, ``Asymmetric
  {{Actor Critic}} for {{Image-Based Robot Learning}}: 14th {{Robotics}}:
  {{Science}} and {{Systems}}, {{RSS}} 2018.'' [Online]. Available:
  \url{http://www.scopus.com/inward/record.url?scp=85127903841&partnerID=8YFLogxK}
\BIBentrySTDinterwordspacing

\bibitem{lengyelVolumetricHierarchical2016}
``Volumetric {{Hierarchical Approximate Convex Decomposition}},'' in \emph{Game
  {{Engine Gems}} 3}, E.~Lengyel, Ed.\hskip 1em plus 0.5em minus 0.4em\relax {A
  K Peters/CRC Press}.

\end{thebibliography}

\end{document}